\documentclass[sigconf]{acmart}
\usepackage{subcaption}
\usepackage{caption}
\usepackage[]{graphicx}
\usepackage{fnpct}
\renewcommand\footnotetextcopyrightpermission[1]{}
\usepackage{amsmath}       
\usepackage{graphicx}     
\usepackage{hyperref}  
\usepackage{footmisc}
\usepackage{url}  
\usepackage{xcolor}
\usepackage{colortbl}
\usepackage{hyperref}
\hypersetup{
    colorlinks=true,
    citecolor=blue,  
    linkcolor=red,   
    urlcolor=cyan    
}
\usepackage[skip=0.5pt]{caption} 
\usepackage{url}
\usepackage{cleveref}

\crefformat{section}{\S#2#1#3} 
\crefformat{subsection}{\S#2#1#3}
\crefformat{subsubsection}{\S#2#1#3}

\AtBeginDocument{%
  \providecommand\BibTeX{{%
    \normalfont B\kern-0.5em{\scshape i\kern-0.25em b}\kern-0.8em\TeX}}}

\title{Do LLMs Exhibit Human-Like Reasoning? Evaluating Theory of Mind in LLMs for Open-Ended Responses}

\author{Maryam Amirizaniani}
\affiliation{
  \institution{University of Washington}
  \city{Seattle}
  \country{USA}
}
\email{amaryam@uw.edu}

\author{Elias Martin}
\affiliation{
  \institution{University of Washington - Bothell}
  \city{Bothell}
  \country{USA}
}
\email{eamart34@uw.edu}

\author{Maryna Sivachenko}
\affiliation{
  \institution{University of Washington - Bothell}
  \city{Bothell}
  \country{USA}
}
\email{msiva@uw.edu}

\author{Afra Mashhadi}
\affiliation{
  \institution{University of Washington - Bothell}
  \city{Bothell}
  \country{USA}
}
\email{mashhadi@uw.edu}

\author{Chirag Shah}
\affiliation{
  \institution{University of Washington}
  \city{Seattle}
  \country{USA}
}
\email{chirags@uw.edu}

\usepackage{bibentry}

\begin{document}
\begin{abstract}

Theory of Mind (ToM) reasoning entails recognizing that other individuals possess their own intentions, emotions, and thoughts, which is vital for guiding one's own thought processes. Although large language models (LLMs) excel in tasks such as summarization, question answering, and translation, they still face challenges with ToM reasoning, especially in open-ended questions. Despite advancements, the extent to which LLMs truly understand ToM reasoning and how closely it aligns with human ToM reasoning remains inadequately explored in open-ended scenarios. Motivated by this gap, we assess the abilities of LLMs to perceive and integrate human intentions and emotions into their ToM reasoning processes within open-ended questions. Our study utilizes posts from Reddit's ChangeMyView platform, which demands nuanced social reasoning to craft persuasive responses. Our analysis, comparing semantic similarity and lexical overlap metrics between responses generated by humans and LLMs, reveals clear disparities in ToM reasoning capabilities in open-ended questions, with even the most advanced models showing notable limitations. To enhance LLM capabilities, we implement a prompt tuning method that incorporates human intentions and emotions, resulting in improvements in ToM reasoning performance. However, despite these improvements, the enhancement still falls short of fully achieving human-like reasoning. This research highlights the deficiencies in LLMs' social reasoning and demonstrates how integrating human intentions and emotions can boost their effectiveness.

\end{abstract}

\ccsdesc[500]{Generative AI~Large Language Models}
\ccsdesc[500]{Generative AI~Information retrieval}
\ccsdesc[500]{Generative AI~Reasoning in LLMs}

\keywords{Generative AI, Large Language Models, Information Retrieval, Reasoning in LLMs}
\maketitle
\pagestyle{plain}

\section{Introduction}

In the realm of artificial intelligence, the ability of machines to understand and respond to human-like social cues—like theory of mind (ToM) reasoning—remains a pivotal challenge. ToM entails the capacity to attribute mental states, such as intention, emotion, and belief, to oneself and others, and to understand that these states can be different from one's own~\citep{leslie2004core}. This cognitive ability is fundamental in social human interaction and crucial for effective communication. Large language models (LLMs), which have achieved impressive success in various natural language processing tasks \cite{shen2023shaping, huang2023chatgpt, yu2023leveraging}, are now being pushed to the frontier of social reasoning to see if they can mimic this quintessentially human trait, especially in interacting with with human.

Despite LLMs' prowess in handling structured tasks such as summarization~\cite{ahmed2022few, kolagar2024aligning, tang2023evaluating}, question answering~\cite{abbasiantaeb2024let, tan2023can}, and language translation~\cite{lu-lin-2023-characterised, kocmi2023findings}, they have shown limitations when tasked with zero-shot ToM reasoning that requires nuanced understanding and integration of human mental state~\cite{ma-etal-2023-towards-holistic, zhou-etal-2023-cast, street2024llm, strickland2023ai}. Many studies have shown this limitation of LLMs via utilizing multiple choice and short answer questions~\cite{wu-etal-2023-hi, kosinski2023theory,kim-etal-2023-fantom}, but not on open-ended questions. We aim to bridge this gap by rigorously evaluating the ability of LLMs to engage in zero-shot ToM reasoning tasks within open-ended scenarios, assessing how closely their performance aligns with human capabilities in ToM reasoning task. In particular, we aim to answer the following research questions:

\begin{itemize}
    \item \textbf{RQ1}: To what degree are LLMs capable of zero-shot reasoning in open-ended questions?
    \item \textbf{RQ2}: To what extent are human and LLM social reasoning capabilities aligned in addressing open-ended questions?
    \item \textbf{RQ3}: How does considering human mental state affect the performance of LLMs in ToM reasoning of open-ended questions?
\end{itemize}

To address the posed research questions, we utilized data from Reddit's ChangeMyView subreddit—a platform noted for its intense social interactions and open-ended questions requiring robust reasoning to shift the viewpoint's of posters. Our study examines the capability of LLMs to provide reasoning and respond effectively to these queries, specifically focusing on how integrating individuals' mental states can enhance this ability.

Through comparative analyses assessing semantic similarity and lexical overlap scores between human and LLM responses, we observed significant disparities in reasoning capabilities within open-ended scenarios, similar to those in non-open-ended questions. These findings reveal considerable limitations in even the most advanced LLMs as mentioned in ~\cite{tunstall2023zephyr} and Huggingface leaderboard~\footnote{\url{https://huggingface.co/spaces/HuggingFaceH4/open_llm_leaderboard}}, such as Zephyr-7B \cite{tunstall2023zephyr}, Llama2-Chat-13B \cite{touvron2023llama}, and GPT-4 \cite{achiam2023gpt}. Our research underscores the effectiveness of incorporating mental states such as human intentions and emotions into LLM reasoning via prompt tuning. The results emphasize the need for LLMs to understand human-like mental states to effectively engage in social reasoning in open-ended questions, indicating a key area for development.

The \textbf{motivation} stems from assessing how well LLMs generate responses to open-ended questions and how their reasoning aligns with human reasoning. This analysis is crucial for understanding LLMs' limitations in replicating human ToM reasoning. The \textbf{novelty} of this research lies in its focus on comparing LLM reasoning with human reasoning in open-ended questions, using Reddit posts as a rich data source. Our \textbf{contributions} to this field are outlined as follows:

\begin{itemize}
    \item We provide a comparative analysis of responses from humans and LLMs, emphasizing the disparities in ToM reasoning capabilities within open-ended scenarios.
    \item We rigorously assess LLMs' reasoning capabilities to integrate human intentions and emotions within open-ended questions, utilizing data from Reddit's ChangeMyView.
    \item We utilize a prompt tuning method that substantially improves the ToM reasoning performance of LLMs in open-ended questions by incorporating human intentions and emotions.
\end{itemize}

The rest this paper is organized as follows: Section \cref{related} reviews related work in the field. Section \cref{Experiment} details the experimental setup and methodology employed in this study. Section \cref{Analysis} is devoted to discussing the analysis and results of the experiment. Section \cref{Discussion} presents a discussion of the findings relative to the experiment. Finally, Sections \cref{Conslusion} and \cref{limitation} concludes the paper by summarizing the key findings and explores the limitations of this study, highlighting areas for future research.

\section{Related Work} \label{related}

This section examines the development of ToM reasoning capabilities in humans and investigates how these principles are being adapted and evaluated within the context of LLMs. Subsequently, we will delve into studies that incorporate the human state of mind in ToM reasoning.

\subsection{Theory of Mind (ToM) Reasoning}

ToM in reasoning refers to the ability to understand and attribute mental states, such as beliefs and intentions, to oneself and others to predict and interpret behavior in social interactions~\citep{leslie2004core}. In exploring the developmental trajectory ToM abilities,~\citet{van-duijn-etal-2023-theory} examined the performance of 11 state-of-the-art language models alongside children aged 7-10 on advanced ToM tests. Notably, instruction-tuned LLMs from the GPT family demonstrated superior performance, occasionally surpassing that of children. However, basic LLMs faced notable challenges in solving ToM tasks~\cite{van-duijn-etal-2023-theory}. In this context, researchers have utilized a variety of cognitive science tests to probe the emergence of ToM reasoning within LLMs~\cite{nematzadeh-etal-2018-evaluating, pu2020program, fried-etal-2023-pragmatics}. These studies aim to determine how well LLMs can emulate the human ability to understand.

Findings from~\citet{sap-etal-2022-neural} using the TOMI dataset~\cite{le-etal-2019-revisiting} indicate that GPT-3, exhibits ToM capabilities that are significantly inferior to those of humans. This gap underscores a critical limitation in current LLMs with some traditional ToM tasks, they continue to struggle with reliably showing ToM capabilities~\cite{ ullman2023large, yu-etal-2023-alert}. These studies suggest that the consistency with which LLMs demonstrate ToM abilities remains questionable, often defaulting to surface-level reasoning strategies rather than engaging in deep, robust ToM reasoning~\cite{shapira-etal-2024-clever}. Moreover,~\citet{kim-etal-2023-fantom} introduced FANTOM benchmark to rigorously assess ToM abilities within conversational contexts. This benchmark has revealed significant challenges facing state-of-the-art LLMs, such as GPT-4, Llama 2, Falcon, and Mistral, particularly in maintaining performance in ToM reasoning tasks in comparison to humans, even with chain-of-thought reasoning or fine-tuning.

~\citet{wu-etal-2023-hi} highlights a significant decline in performance across various LLMs, such as GPT 4, GPT 3.5, Claude, and Guanaco, when tasked with higher-order ToM challenges. This trend suggests that as the complexity of ToM tasks increases, the ability of these models to accurately interpret and respond to nuanced mental states diminishes notably.

On the other hand, these findings contradict the claims by~\citet{kosinski2023theory} and reiterated by~\citet{bubeck2023sparks}, which posited that modern LLMs reached high ToM scores.

\subsection{Mental States in ToM}

Recent research in the application of ToM within LLMs has provided notable insights into the effects of understanding mental states on LLM reasoning. For instance, the Foresee and Reflect (FaR) framework offers a reasoning structure that encourages LLMs to anticipate future challenges and reason about potential actions. Analysis reveals the effectiveness of incorporating mental states into reasoning~\cite{zhou2023far}. Additionally,~\citet{ma-etal-2023-towards-holistic} develops a comprehensive taxonomy for ToM in LLMs, known as Abilities in Theory of Mind Space (ATOMS), which categorizes crucial components such as Intentions, Percepts, Beliefs, Emotions, Knowledge, Desires, and Non-literal Communication. This framework aims to provide a structured approach to assess and systematically enhance ToM capabilities. These developments highlight the effectiveness of recent model iterations in approximating human mental states.

Concurrently, the BigToM benchmark has been developed to specifically assess LLMs' social reasoning capabilities, focusing on aspects such as beliefs, percepts, desires, and user actions~\cite{gandhi2024understanding}. Another tool that has emerged mental state in ToM is SymbolicToM, which enhances ToM capabilities in reading comprehension tasks by effectively representing entities' beliefs and facilitating higher-order reasoning. This approach has shown promise in providing a deeper understanding of belief states and their implications for ToM~\cite{sclar-etal-2023-minding}.

Moreover, different methods are utilized in ToM reasoning understanding. For instance,~\citet{li-etal-2023-theory} focused on ToM into dialogue models through reinforcement learning have demonstrated significant improvements in the quality of guidance these systems provide. This integration underscores the importance of accurately interpreting and responding to intents within conversational contexts~\cite{li-etal-2023-theory}. Additionally,~\citet{lim2020improving} explores the integration of the ``Bayesian Theory of Mind'' with optimal-planning agents. This approach illustrates how explicitly representing others' intentions can enhance performance in ToM reasoning.

In contrast to earlier research, which typically relied on multiple-choice formats or short-answer questions to evaluate the ToM capabilities of LLMs~\cite{wu-etal-2023-hi}, our study adopts a more nuanced approach by utilizing open-ended questions. This method allows for a broader range of responses, providing deeper insights into how LLMs interpret and respond to complex scenarios. By shifting from structured to more exploratory questioning, we aim to uncover subtler aspects of ToM reasoning in LLMs.

\section{Experimental Setup}
\label{Experiment}
In this section, we provide details of the LLMs and the datasets used for our experiments.

\subsection{Large Language Models}

For the work reported here, we utilize three of the most popular LLMs known for their exceptional reasoning capabilities, as documented in~\cite{tunstall2023zephyr}. Specifically, we have selected Zephyr-7B~\cite{tunstall2023zephyr}, Llama2-Chat-13B~\cite{touvron2023llama}, and GPT-4 ~\cite{achiam2023gpt}, setting the temperature parameter to 0.5.

\subsection{Dataset}

For our experiments, we leverage the Reddit r/ChangeMyView (CMV) community\footnote{\url{https://www.reddit.com/r/changemyview/}} as a source for prompts, capitalizing on its reputation for open-ended discussions. This platform is ideal as it enables questioners to share opinions and invite others to change these views with reasoning grounded in the questioners' beliefs, thoughts, intent, and other mental states, thus providing a rich dataset for this study.

The data for this study is comprised of three parts. The first part consists of Reddit posts, which were crawled from the aforementioned Reddit community. We initially collected the 1,000 most recent posts using the PRAW Reddit API\footnote{https://www.reddit.com/r/PRAW/}, with the oldest one having been published in February 2024. It is worth noting that these posts could not have been incorporated in the training corpora of the LLMs examined in this study, as their most recent training data does not encompass data from February 2024 and more recent. The other two parts include human-written and LLM-generated responses to those posts, which will be discussed in more detail subsequently.

For the human-written responses, we extracted the top five responses based on community upvotes, utilizing Reddit's internal voting mechanism to rank comments. On Reddit, the `Vote Number' of a post or comment represents its net score, calculated by subtracting the number of downvotes from the number of upvotes. To ensure reliability in our analysis, we selected only those responses with entirely positive scores. By selecting the top five responses, we ensure a diverse and well-supported dataset of argumentative responses, providing a comprehensive view of human reasoning. After filtering out data with fewer than five responses and further cleaning and preprocessing to ensure all columns were non-null and formatted as strings, we retained 845 distinct Reddit posts, each with five Reddit user responses.

Simultaneously, to create a dataset of LLM-generated responses, we input the same 845 posts into the above-mentioned LLMs using a specially crafted prompt template and requested it to generate five responses for each.

To develop this prompt template, we conducted several iterations. Initially, we utilized a generic prompt template, \texttt{``Generate five reasoning answers for `Question'.}'', which failed to produce satisfactory results. The ineffectiveness of this prompt template is attributed to its lack of specificity and inherent ambiguity, leading LLMs to generate generic and irrelevant responses rather than precise and relevant ones.

\begin{figure}[h]
    \centering
    \includegraphics[scale=0.43]{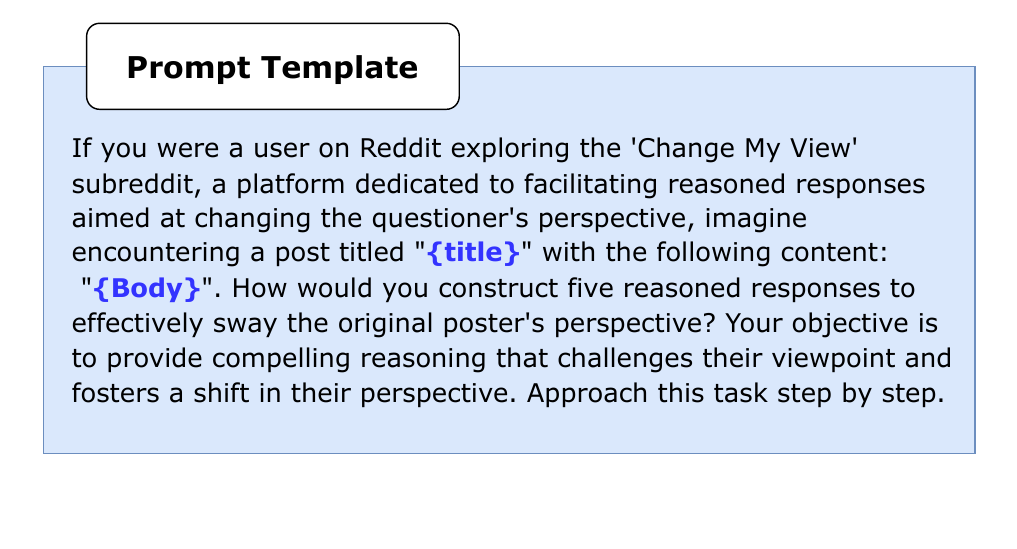}
    \caption{Initial prompt template for generating reasoning answers.}
    \Description{This is a descriptive text}
      \vspace{-0.5em}
    \label{fig:template}
\end{figure}

To address this challenge, we were inspired by the Chain of Thought (CoT) approach \cite{wei2022chain}, which promotes a step-by-step reasoning process in LLMs, resulting in higher-quality answers~\cite{liu2023federated, ranaldi-freitas-2024-aligning, wang2023cuecot, kim-etal-2023-fantom}. Additionally, following \cite{kojima2022large, miao2024selfcheck}, we enhanced our approach by clearly defining the tasks and setting explicit expectations within the prompts. This adjustment has been shown to significantly improve the quality of responses as supported by \citet{ye2024investigating} and \citet{zeng2024evaluating}, thus ensuring that our prompts are well-crafted to elicit detailed and contextually appropriate answers from the models. These comprehensive enhancements led to the development of the final prompt template, which not only elicited straightforward answers but encouraged reasoning that challenges and potentially shifts questioner viewpoints, as shown in Figure~\ref{fig:template}. This approach mirrors the nature of the conversation on the Reddit CMV platform.

\subsection{Data Description}
\begin{figure*}[ht]
    \centering
    \begin{subfigure}[b]{0.22\textwidth}
        \centering
        \includegraphics[width=\textwidth]{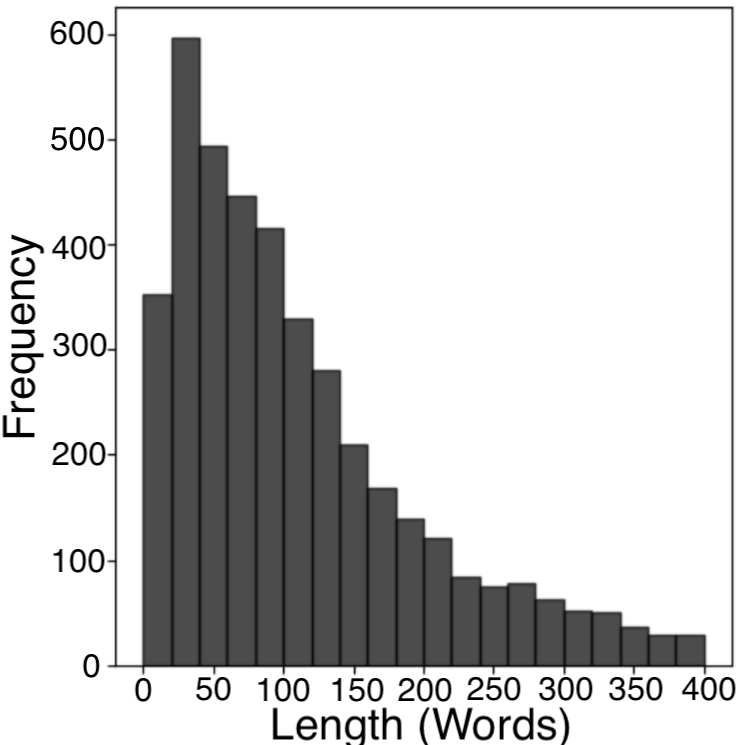}
        \caption{Human comments}
        \label{fig:first_image}
    \end{subfigure}
    \hfill 
    \begin{subfigure}[b]{0.22\textwidth}
        \centering
        \includegraphics[width=\textwidth]{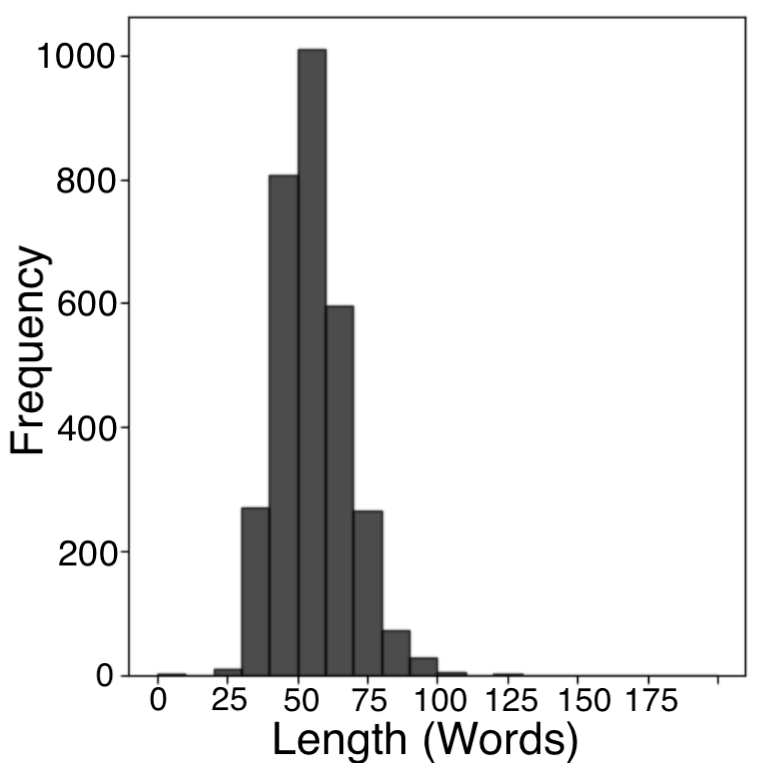}
        \caption{GPT-4 comments}
        \label{fig:second_image}
    \end{subfigure}
    \hfill 
    \begin{subfigure}[b]{0.22\textwidth}
        \centering
        \includegraphics[width=\textwidth]{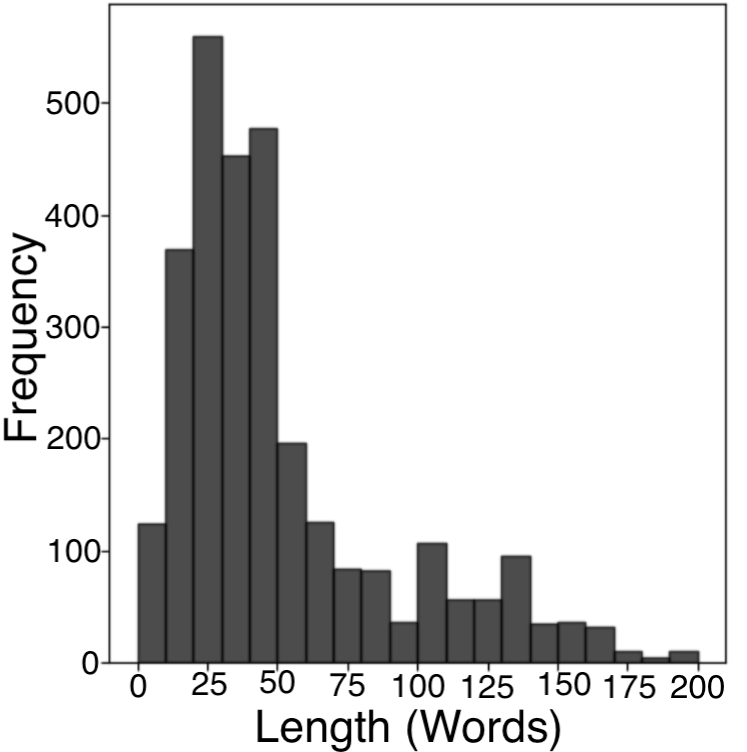}
        \caption{Llama2-Chat-13B comments}
        \label{fig:third_image}
    \end{subfigure}
    \hfill 
    \begin{subfigure}[b]{0.22\textwidth}
        \centering
        \includegraphics[width=\textwidth]{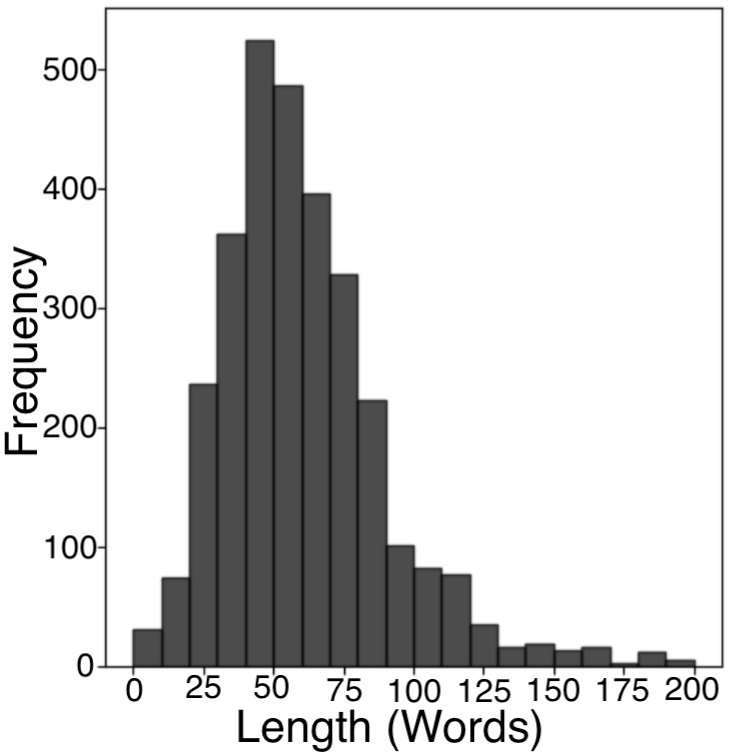}
        \caption{Zephyr-7B comments}
        \label{fig:fourth_image}
    \end{subfigure}
    \caption{Overview of length distribution of reasoning answers generated by humans and LLMs.}
    \Description{This is a descriptive text for the example image}
    \label{fig:overall_visualizations}
\end{figure*}

Drawing on the dataset mentioned above, Figure \ref{fig:overall_visualizations} illustrates the distribution of average lengths for comments generated by both humans and LLMs. This visualization effectively highlights the variability in response lengths from each source, with each figure presenting a distribution that approximates a normal curve. Notably, human responses range up to 400 words but typically they are around 50 words. Also, responses from GPT-4.0 tend to cluster around an average of 50 words, demonstrating a narrower range in length variability. Conversely, Llama2-Chat-13B and Zephyr-7B display a broader spectrum in the length of their responses, suggesting a richer diversity in response detail and complexity.

Despite the numerical differences in length, the overall patterns observed across the distributions underscore an underlying consistency in how different models and humans generate response lengths.

\section{Analysis and Result} \label{Analysis}

In this section, we delve into a detailed exploration and analysis of the data we  collected. To ensure a thorough evaluation, we  employed two distinct approaches: human-based evaluation, which relies on assessments from human evaluators, and metric-based evaluation, which utilizes NLP quantitative evaluation metrics. Below, we will present a comprehensive report detailing the analysis and results derived from each of these approaches.

\subsection{\textbf{Human-based Evaluation}}

We utilize the Human-in-the-Loop (HIL) method to evaluate the reasoning capabilities of LLMs. To ensure a comprehensive assessment, trained assessors, who are graduate students in computer science, were asked to annotate the responses generated by these models.

The assessment process is divided into two rounds to comprehensively analyze the reasoning abilities of LLMs. In the initial round, we evaluate the LLMs’ reasoning performance using the previously mentioned prompt template, aiming to gauge each model's basic reasoning capabilities. The second round intensifies the scrutiny by incorporating details of mental states, such as questioners' intentions, emotions, and sentiments embedded within the questions. The LLMs are tasked with generating reasoning responses that consider these aspects of mental states. Our annotators then evaluate these responses to determine the quality of reasoning, with focusing on how the integration of mental state details influences the reasoning process.

In this phase, we randomly selected 100 posts from our dataset. Each post generated five reasoning answers, resulting in a total of 500 answers per LLM. With three LLMs under evaluation, this led to an aggregate of 1,500 reasoning answers for human assessment. To facilitate the labeling process, we recruited and trained ten assessors, ensuring they were well-versed in the goal and the task, while not having any conflict that could jeopardize this work. Each reasoning answer produced by the LLMs was independently evaluated twice by different assessors. The same set of questions was employed for labeling in both the first and second rounds. However, we ensured that assessors did not encounter the same questions in the second round to mitigate any familiarity bias.

For assessment, it is essential to establish specific criteria for assessors to evaluate the reasoning in responses across these two rounds of assessment. Drawing from previous research on the `Reasoning' task, we focus on the criterion of ``Reasoning Correctness'', which mandates that the reasoning in responses should be valid, supporting the conclusions or explanations logically and accurately~\cite{prasad-etal-2023-receval, ling2023deductive}. The criteria for `Reasoning Correctness' are detailed as follows: 

\begin{itemize}
    \item \textbf{Reasoning Correctness}: Does the reasoning in the response adequately address the question? Is the reasoning relevant to the question asked? Does the reasoning accurately represent the information provided and the conclusions drawn from it? Is the reasoning logically sound, without any fallacies or errors in logic? Does the reasoning lead to a clear and well-supported conclusion?
\end{itemize}

To ensure consistency in our evaluation process, we calculate the Agreement Rate among assessors to quantify the level of consensus on evaluation results. We adopt established statistical methods to determine inter-assessor agreement, primarily using Cohen's kappa~\citep{10.1162/coli.07-034-R2}, considered acceptable within the range of 0.61 to 0.80~\citep{10.1162/coli.07-034-R2}. Additionally, we measure the overlap rate among the responses provided by assessors, with an acceptable benchmark set at 80\%. These methods provide a rigorous and quantifiable measure of inter-assessor agreement, thereby enhancing the reliability of our evaluations. The subsequent sections will detail the methodologies employed in the two rounds of assessments and discuss the results.

\subsubsection {\textbf{Human annotation - Round 1}}

This section outlines the first round of human annotation used to assess the reasoning capabilities of LLMs. We initially implemented a 5-point Likert scale to evaluate responses based on the "Reasoning Correctness" criteria but encountered significant challenges. The Cohen Kappa scores, as shown in Table~\ref{tab:agreement-rate}, indicated that this method did not fulfill the acceptance criteria for consistent inter-annotator agreement. As a result, we transitioned to a 3-point Likert scale, which yielded a Cohen’s kappa score of 0.4751 and overlap rate of 61.73\%. Although these scores represent an improvement, they still fail to achieve satisfactory agreement. This may have been due to the diversity of reasoning approaches among the annotators. In response, further iterations of our study introduced a dichotomous scale, which enhanced the Cohen’s kappa score to 0.6618 and overlap rate to 83.66\%, thereby meeting the agreement standards. This adjustment effectively reduced subjectivity among assessors in evaluating the reasoning answers.

Our findings reveal variation in the reasoning quality of responses to open-ended questions: 23.52\% from Zephyr-7B, 28.19\% from Llama2-Chat-13B, and 35.69\% from GPT-4 were rated as demonstrating adequate reasoning, according to the consensus among assessors. These results are detailed in Table~\ref{tab:Labeling-result}. This analysis indicates that from a human perspective, LLMs struggle to generate reasonable and contextually appropriate responses to open-ended questions.

\begin{table}[htbp]
\centering
\footnotesize
\caption{Agreement rate among annotators for ``Reasoning Correctness'' in the first and second rounds.}
\label{tab:agreement-rate}
\begin{tabular}{l c c}
\hline
\toprule
Scale & Cohen’s kappa score  & Overlap rate\\
\hline
\rowcolor{gray!20}
\multicolumn{3}{c}{\textbf{Round 1}} \\
\hline
5-point Likert scale & 0.3527 & 45.10\%\\
3-point Likert scale & 0.4751 & 61.73\%\\
Dichotomous scale & 0.6618 & 83.66\% \\
\hline
\rowcolor{gray!20}
\multicolumn{3}{c}{\textbf{Round 2}} \\
\hline
Dichotomous scale & 0.7204 & 86.24\% \\
\bottomrule
\end{tabular}
\end{table}

\begin{table}[htbp]
\centering
\footnotesize
\caption{Assessor `Yes' responses to the `Reasoning Correctness' criterion across the first and second rounds, and delta change of these two rounds.}
\label{tab:Labeling-result}
\begin{tabular}{lc}
\hline
\toprule
Metric & Reasoning Quality = "Yes" \\
\hline
\rowcolor{gray!20}
\multicolumn{2}{c}{\textbf{Round 1}} \\
\hline
Zephyr-7B & 23.52\% \\
Llama2-Chat-13B & 28.19\% \\
GPT-4 & 35.69\% \\
\hline
\rowcolor{gray!20}
\multicolumn{2}{c}{\textbf{Round 2}} \\
\hline
Zephyr-7B & 29.43\% \\  
Llama2-Chat-13B & 35.09\% \\  
GPT-4 & 43.21\% \\ 
\hline
\rowcolor{gray!20}
\multicolumn{2}{c}{\textbf{Delta improvement}} \\
\hline
Zephyr-7B & $\uparrow 5.91\%$ \\  
Llama2-Chat-13B & $\uparrow 6.9\%$\\  
GPT-4 & $\uparrow 7.52\%$ \\  
\bottomrule
\end{tabular}
\end{table}


\subsubsection{\textbf{Human annotation - Round 2}} \label{sec:Human}

\begin{figure}[h]
    \centering
    \includegraphics[scale=0.45]{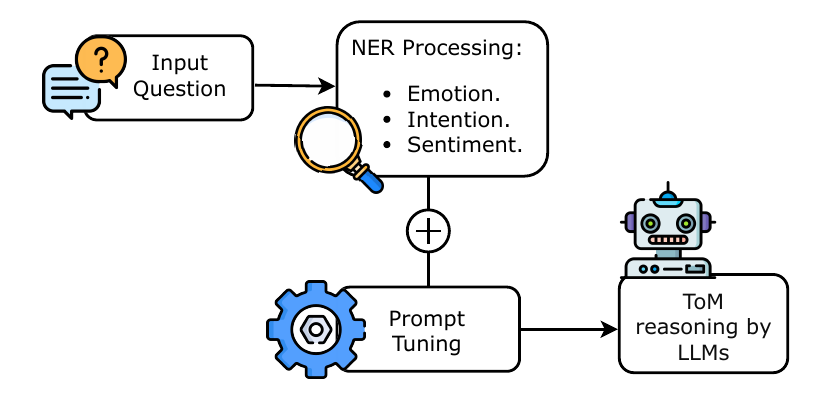}
    \caption{ToM reasoning process via prompt tuning.}
    \Description{This is a descriptive text for the example image}
      \vspace{-0.5em}
    \label{fig:flow}
\end{figure}

In this evaluation round, which includes human annotation, we investigate how detailed mental states—such as emotions, intentions, and question sentiment—affect the reasoning performance of LLMs. Following the findings of~\citet{kim-etal-2023-fantom}, which highlight the effectiveness of prompt tuning for integrating mental states over other methods, we have incorporated these elements into our analysis by embedding them into prompts. This prompt tuning approach subsequently leads LLMs to generate reasoning responses based on these enriched prompts. The process of this integration is depicted in Figure~\ref{fig:flow}. We initiate this phase by extracting the mental states information from the dataset. Subsequent sections will provide a detailed exposition of the methodologies utilized for the extraction of mental states.

\begin{itemize}
\item \textbf{Posts sentiment and emotion:} To measure the sentiment and emotions of the posts, we utilized a RoBERTa~\cite{liu2019roberta} base model that had been fine-tuned on the Go Emotions dataset~\cite{demszky-etal-2020-goemotions}, which is derived from Reddit posts. This choice was ideal for our study since Go Emotions can predict and rank a range of up to 28 distinct emotions based on the likelihood of their association with a given text input. For our analysis, we concentrated on the highest-ranked emotion identified in both the titles and bodies of the posts. Furthermore, to precisely capture and analyze the spectrum of emotions conveyed by questioners in their inquiries, we utilized Named Entity Recognition (NER) techniques. The NER is an information extraction technique that identifies and categorizes named entities in text into predefined categories. Here, we employed NER to specifically identify and classify emotional expressions within text, enhancing our understanding of the distinct emotional states expressed by questioners.

Additionally, we categorized these 28 emotions into three sentiment groups—positive, negative, or neutral—based on which sentiment classification aligned most closely with each emotion, following the methodologies outlined in~\cite{prajapati2023analysis}.

To better understand the emotions and sentiments in the posts and responses generated by both humans and LLMs, we conducted a series of analyses, with results presented in Table~\ref{tab:Stat-Sig-Emotion-Sentiment}. We used a pairwise t-test to assess the statistical significance of differences in sentiments and emotions, considering p-values of 0.05 or lower as significant. Notably, human responses showed a broader emotional range compared to LLMs, which tended to express concentrated emotions, primarily admiration, approval, neutrality, and curiosity. It's also important to highlight that the Zephyr-7B LLM exhibited a notably limited emotional range, with significant findings in only one emotion, suggesting it is less emotionally developed than more advanced models like GPT-4 and Llama2-Chat-13B.

\begin{table}
\centering
\footnotesize
\caption{Emotion sentiment significance results for top emotions in reasoning answers generated by Human, GPT-4, Llama2-Chat-13B, and Zephyr-7B. Statistically significant values are shown in \textbf{bold}. }
\label{tab:Stat-Sig-Emotion-Sentiment}
\begin{tabular}{l c c c }
\hline
\toprule
  Emotion & Positive & Negative & Neutral  \\
 \hline
\rowcolor{gray!20}
\multicolumn{4}{c}{\textbf{Human}} \\
\hline
Admiration & \textbf{+6.42e-08} & 2.53e-01 & 5.26e-02\\
Caring & 2.58e-01 & \textbf{+2.44e-04} & 2.01e-01\\
Approval & \textbf{+1.56e-02} & 9.54e-01 & 2.52e-01\\
Neutral & \textbf{-2.51e-02} & 3.43e-01 & 1.33e-01\\
Desire & 8.72e-01 & \textbf{+1.43e-03} & 1.36e-01\\
Disappointment & 6.01e-01 & \textbf{+1.82e-02} & 1.61e-01\\
Anger & \textbf{+2.28e-05} & 5.42e-01 & 9.73e-02\\
Sadness & 7.63e-01 & \textbf{+1.87e-02} & 1.95e-01\\
\hline
\rowcolor{gray!20}
\multicolumn{4}{c}{\textbf{GPT-4}} \\
\hline
Admiration & \textbf{+2.13e-04} & 1.26e-01 & 3.22e-01 \\
Caring & 4.01e-01 & \textbf{+2.20e-02} & 1.47e-01\\
Approval & \textbf{+4.52e-02} & \textbf{-2.73e-02} & 9.16e-01\\
Neutral & \textbf{-7.67e-03} & 4.64e-01 & 3.77e-01\\
Curiosity & 6.02e-02 & \textbf{+4.56e-02} & 7.34e-02\\
\hline
\rowcolor{gray!20}
\multicolumn{4}{c}{\textbf{Llama2-Chat-13B}} \\
\hline
Joy & \textbf{+3.78e-03} & 6.94e-01 & 2.52e-01 \\
Admiration & \textbf{+4.60e-14} & 1.10e-01 & \textbf{-6.46e-03}\\
Approval & \textbf{+3.31e-04} & 1.06e-01 & 3.66e-01\\
Neutral & \textbf{-5.77e-04} & 9.57e-01 & 1.21e-01\\
Curiosity & \textbf{+1.16e-05} & 9.55e-01 & \textbf{-4.19e-02}\\
Annoyance & 2.48e-01 & \textbf{+1.11e-16} & \textbf{-9.97e-04}\\
\hline
\rowcolor{gray!20}
\multicolumn{4}{c}{\textbf{Zephyr-7B}} \\
\hline
Joy & 4.11e-01 & \textbf{+7.41e-03} & 1.05e-01 \\ 
\hline
\toprule
\end{tabular}
\end{table}

\item \textbf{Posts intention:} To measure the intentions behind the posts, we utilized a DistilBERT model~\cite{sanh2019distilbert} fine-tuned on the Clinc OOS dataset~\cite{larson-etal-2019-evaluation}. This dataset includes 150 intent classifications designed for task-oriented dialogue systems, sourced from a variety of platforms including Quora and Wikipedia. Similar to how we extract emotions, we also employed NER techniques to identify intentions behind the questions. When the NER identifies intents as `OOS' (Out of Scope), indicating that they do not align with predefined categories, we particularly focus on extracting tokens that carry the highest `OOS' weight. This method enhances our accuracy in identifying intents that fall outside the standard classifications.

\end{itemize}

\begin{figure}
    \centering
    \includegraphics[scale=0.45]{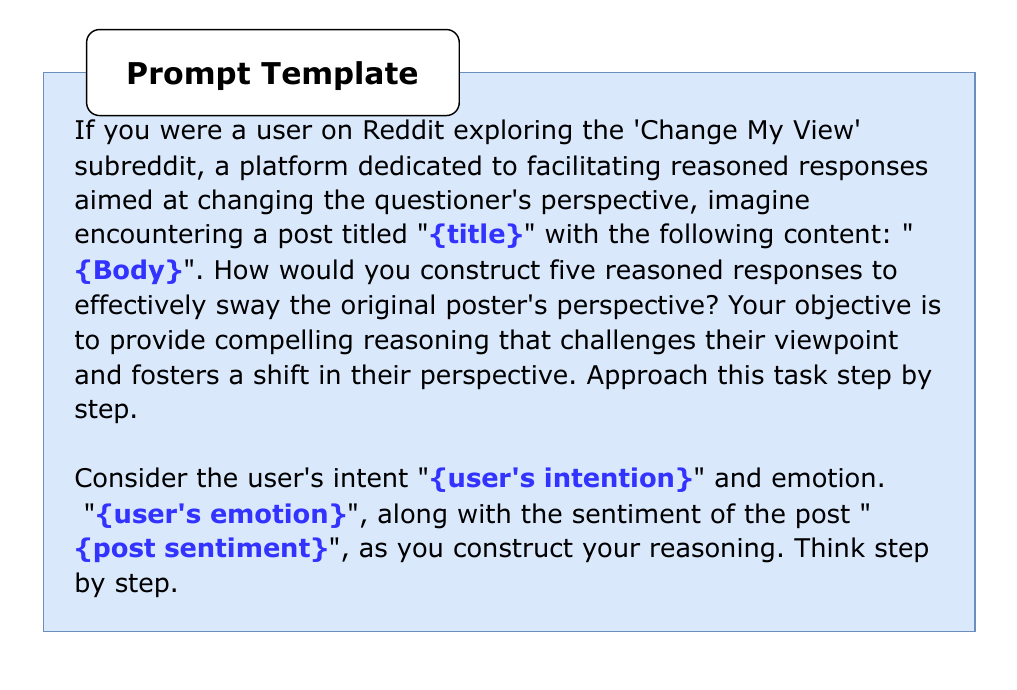}
    \caption{Final prompt template.}
    \Description{Description of the second example image.}
      \vspace{-0.5em}
    \label{fig:template2}
\end{figure}

After extracting sentiment, emotion, and intention from questions, we refined the prompt by incorporating this contextual data. According to \cite{lin2024write}, such contextual information enhances output quality. The updated prompt template is illustrated in Figure~\ref{fig:template2}. This comprehensive prompt allows for a rich exploration of how these mental states affect model performance in reasoning. We then asked the LLMs to regenerate reasoning responses based on the updated prompt and the given mental states. Annotators were tasked with evaluating the ``Reasoning Correctness'' using the previously defined criteria and scale. In this round, we achieved a Cohen’s kappa score of 0.7204 and an overlap rate of 86.243\% for the Dichotomous scale which met the acceptance rate. The results showed an improvement in reasoning quality: 29.43\% for Zephyr-7B, 35.09\% for Llama2-Chat-13B, and 43.21\% for GPT-4. In the comparative analysis between the first and second rounds, the delta improvements observed were as follows: Zephyr-7B exhibited an increase of 5.91\%, Llama2-Chat-13B showed a gain of 6.9\%, and GPT-4 demonstrated an enhancement of 7.52\%. These findings are presented in Table~\ref{tab:Labeling-result}.

In summary, this evaluation phase involved human annotators assessing the impact of incorporating mental states—like sentiment, emotion, and intent—into LLM reasoning for open-ended questions. Drawing on the results presented in Table~\ref{tab:Labeling-result} and the observed delta improvement, we conclude that while incorporating mental states into reasoning generation enhances the performance of LLMs in producing responses to open-ended questions, these models still fall short of generating high-quality reasoning responses.

\subsection{\textbf{Metric-based Evaluation}}

Human evaluation shows that incorporating mental states improves LLM reasoning in open-ended questions, but the quality remains inadequate. In this section, we aim to quantitatively assess how closely LLM-generated responses mimic the reasoning patterns typically found in human responses. To do this, we conduct a detailed analysis by comparing LLM outputs to human reasoning, using advanced metrics for measuring semantic similarity and lexical overlap, following previous reasoning studies like~\cite{madaan-etal-2022-language, yu-etal-2023-alert, anonymous2024one}.

In our analysis, we use human-written responses as benchmarks to evaluate the outputs of LLMs. For each post, we compare five top-scored human-written answers with five answers generated by the LLMs. We employ evaluation metrics to construct a 5x5 score matrix for each comparison, where each entry represents the comparison between one human answer and one LLM answer. We then apply a max-mapping strategy, matching each LLM-generated answer with the human-written answer that yields the highest score. The average of these maximum scores is calculated to obtain the final scores, providing a detailed quantitative analysis of how closely LLM outputs align with human reasoning. The evaluation metrics used include:

\begin{itemize}

\item \textbf{ROUGE-L}~\citep{lin2004rouge}: ROUGE-L is a conventional metric that assesses F1 N-gram overlaps between a candidate sequence and, ideally, several reference sequences. 

\item \textbf{BLEURT}~\citep{sellam2020bleurt}: BLEURT is a metric developed using BERT to detect nuanced semantic similarities between sentences. 

\item \textbf{BERTScore}~\citep{zhang2019bertscore}: The BERTScore utilizes pre-trained BERT embeddings to calculate the semantic similarity between the generated responses and the reference. Here we use “all-mpnet-base-v2” and cosine similarity to evaluate semantic similarity in responses.

\item \textbf{MoverScore}~\citep{zhao2019moverscore}: MoverScore is a reference-based evaluation metric that utilizes Earth Mover's Distance to compare a candidate sentence with its reference, using contextual word representations.

\end{itemize}

The results of our analysis, summarized in Table~\ref{tab:results-Vanilla}, establish a baseline for this research. Here, GPT-4 emerges as the standout model, consistently outperforming the others across all metrics. It scored 0.279 on ROUGE, 0.411 on BLEURT, 0.593 on BERTScore, and 0.310 on MoverScore, confirming its ability to generate responses that are contextually relevant, aligned with the nuances of human reasoning and emotional characteristics of speech.

The second-best performances varied among the other models: Llama2-Chat-13B showed strong results with the second-highest scores in ROUGE, BERTScore, and MoverScore, reflecting its capability to produce responses with significant lexical and semantic resemblance to human output. Meanwhile, Zephyr-7B claimed the second spot in the BLEURT metric, indicating a reasonable level of semantic understanding, though not as pronounced as GPT-4 or Llama2-Chat-13B.

This layered analysis reveals that while no models currently achieve perfect, high-quality human-like reasoning, GPT-4 demonstrates better performance in generating reasoning responses compared to the other two models. This finding aligns with the outcomes of human assessments, indicating that recent advancements in model training are progressively reducing the cognitive discrepancies between human and machine-generated reasoning.

\begin{table}[htbp]
\footnotesize
\centering
\caption{The baseline results for ROUGE-L, BLEURT, BERTScore, and MoverScore were obtained from various LLMs. The highest scores for each model are highlighted in \textbf{bold}, and the second highest with \underline{underline}.}
\label{tab:results-Vanilla}
\begin{tabular}{l c c c c}
\hline
\toprule
 & Zephyr-7B & Llama2-Chat-13B & GPT-4  \\
\hline
ROUGE-L & 0.240 & \underline{0.244} & \textbf{0.279} \\ 
BLEURT & \underline{0.397} & 0.396 & \textbf{0.411} \\ 
BERTScore & 0.582 & \underline{0.585} & \textbf{0.593} \\
MoverScore & 0.300 & \underline{0.301} & \textbf{0.310}\\
\hline
\toprule
\end{tabular}
\end{table}

To determine whether the differences in performance between LLMs across various evaluation metrics are statistically significant, we employ a pairwise t-test approach. Specifically, we regard the differences as statistically significant if the resulting p-values are less than or equal to 0.05. The p-values for each metric are listed in Table~\ref{tab:T-test}, with statistically significant results marked by asterisks.

Analysis shows significant performance differences between models, particularly in ROUGE-L and BLEURT metrics. Specifically, GPT-4 consistently demonstrates statistically significant differences compared to both Zephyr-7B and Llama2-Chat-13B in ROUGE-L and BLEURT, indicating its better performance in these areas. For the BLEURT metric, the comparison between GPT-4 and Zephyr-7B is highly significant, with a p-value below 0.001, suggesting a robust disparity in BLEURT performance favoring GPT-4. However, in the BERTScore metric, none of the model comparisons reached statistical significance, indicating comparable performance across all models in this metric. Lastly, in the MoverScore metric, significant differences were also observed between GPT-4 and the other two models, with more pronounced significance in the comparisons with Llama2-Chat-13B. These findings underscore GPT-4's generally higher ability to reason across most assessed metrics, except for BERTScore, where all models performed similarly.

\begin{table}[htbp]
\centering
\footnotesize
\caption{T-test comparison of LLMs showing statistically significant differences across metrics, indicated by asterisks.}
\label{tab:T-test}
\begin{tabular}{l c c c }
\hline
\toprule
 & Zephyr-7B & Llama2-Chat-13B & GPT-4  \\
 \hline
\rowcolor{gray!20}
\multicolumn{4}{c}{\textbf{ROUGE-L}} \\
\hline
Zephyr-7B &$-$ & 0.217& 0.011 \text{*}  \\ 
Llama2-Chat-13B & 0.217 &$-$ & 0.026\text{*}\\ 
GPT-4 & 0.011 \text{*} & 0.026\text{*} &$-$ \\
\hline
\rowcolor{gray!20}
\multicolumn{4}{c}{\textbf{BLEURT}} \\
\hline
Zephyr-7B & $-$& 0.926 & 0.0008 \text{***}\\ 
Llama2-Chat-13B & 0.926 &$-$ & 0.019\text{*} \\ 
GPT-4 &0.0008 \text{***} & 0.019\text{*} & $-$\\
\hline
\rowcolor{gray!20}
\multicolumn{4}{c}{\textbf{BERTScore}} \\
\hline
Zephyr-7B & $-$ & 0.631 & 0.402  \\ 
Llama2-Chat-13B & 0.631 &$-$ & 0.812\\ 
GPT-4 & 0.402 & 0.812 &$-$ \\
\hline
\rowcolor{gray!20}
\multicolumn{4}{c}{\textbf{MoverScore}} \\
\hline
Zephyr-7B & $-$& 0.128 & 0.034\text{*}  \\ 
Llama2-Chat-13B & 0.128 &$-$ & 0.009 \text{**} \\ 
GPT-4 & 0.034\text{*} & 0.009 \text{**} & $-$ \\
\hline
\toprule
\multicolumn{4}{r@{}}{p-values codes: 0 ‘***’ 0.001 ‘**’ 0.01 ‘*’ 0.05 }
\end{tabular}
\end{table}

To evaluate the effectiveness of LLMs in understanding individual mental states for ToM reasoning, we systematically tune the prompt template to integrate these mental state dimensions, following the format outlined in Section §\ref{sec:Human}. Initially, we modify the prompt template similarly to the one shown in Figure~\ref{fig:template2}, to reflect the sentiment of the question. Subsequent adjustments were made to accommodate the expressed emotion, and the prompt was refined to align with the underlying intention of the question. Ultimately, we incorporated all three mental states—sentiment, emotion, and intention—into the prompt template (Figure~\ref{fig:template2}), providing a comprehensive assessment of the LLMs' reasoning capabilities. The impact of incorporating these elements is detailed in Table~\ref{tab:Final-results}, with an extensive analysis of these findings provided in the following sections.

\subsubsection{\textbf{LLM reasoning with a focus on ``Sentiment''}:}

To assess how understanding the sentiment of questions affects reasoning capabilities, this phase integrates sentiment information into the prompt template. As described in Section~\cref{sec:Human}, we categorize questions into positive, negative, and neutral sentiments. We then tune the prompt template based on these categories and instruct LLMs to generate reasoning responses accordingly.

In this evaluation, GPT-4 clearly outperforms all other models across various metrics, thereby consolidating its status as the top-performing model in sentiment-based ToM reasoning. GPT-4 consistently excels at interpreting and responding based on the sentiment of posts. However, in MoverScore, the margin of its lead narrows, with GPT-4 scoring 0.316, just slightly ahead of Zephyr-7B's 0.310.  Meanwhile, Llama2-Chat-13B consistently ranks as the second-best performer in these metrics, although it falls behind in MoverScore. This indicates that while Llama2-Chat-13B is reliable, it does not lead the field in reasoning among the models tested.

\begin{figure}[h]
    \centering
    \includegraphics[width=0.35\textwidth]{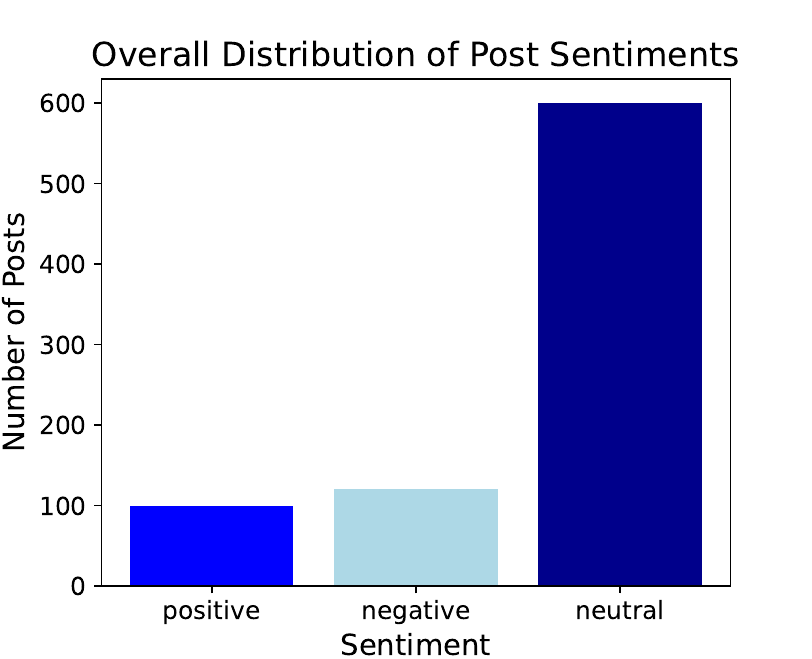}
    \caption{The overall distribution of sentiments of the posts.}
    \Description{Description of the second example image.}
    \label{fig:sentiment}
\end{figure}  

When comparing the evaluation metrics of the sentiment-based reasoning output to the original reasoning output (Table~\ref{tab:Final-results}), we do not observe significant differences. To investigate further, we analyzed the sentiment of the posts, as shown in Figure~\ref{fig:sentiment}. The lack of improvement could be due to the predominance of `neutral' classifications. The overwhelming presence of neutral sentiments in the dataset may dilute the impact of sentiment-based enhancements, as neutral sentiments typically provide fewer distinct cues for the model to leverage in its reasoning processes.

\subsubsection{\textbf{LLM reasoning with a focus on ``Emotion''}:}

This step involves incorporating questioners' emotions into the prompt for reasoning generation, based on the extracted NER data. Similar to our analysis of sentiment, we tune the prompts according to the extracted emotions and instruct the LLMs to generate reasoning responses accordingly. 

As shown in Table~\ref{tab:Final-results}, in the realm of emotion-based tuning, GPT-4 continues to perform better than the other two LLM models, scoring the highest across all evaluation metrics. Notably, it achieves a ROUGE-L score of 0.291 and a BLEURT score of 0.456, underscoring its adeptness at integrating emotional context into its reasoning. The competition, particularly from Llama2-Chat-13B, remains intense, with close scores such as 0.286 in ROUGE-L and 0.441 in BLEURT, reflecting a tight race in model performance when emotion is a significant factor. These scores suggest that all models exhibit improved  ToM reasoning when they integrate emotional information.

\subsubsection{\textbf{LLM reasoning with a focus on ``Intention''}:}

To explore the effect of understanding the intentions behind questions on the reasoning capabilities of LLMs, we incorporate these intentions into the prompt template for reasoning generation, using intentions extracted via NER techniques. This step involves prompt tuning based on these intentions, guiding LLMs to generate responses that consider the specified intent.

Based on the result it is evident that GPT-4 outperforms other models, achieving the highest scores across all evaluation metrics: ROUGE-L at 0.300, BLEURT at 0.452, BERTScore at 0.631, and MoverScore at 0.368. Notably, Zephyr-7B achieves the second-highest score in ROUGE-L, while Llama2-Chat-13B secures the second position in the remaining metrics.

By comparing these outcomes with the performance metrics from the original model of reasoning (Table~\ref{tab:Final-results}), a clear trend of improvement is observable across all metrics for all LLMs tested. This improvement underscores the significant impact that incorporating the explicit intention into the prompts has on the LLMs' ability to approximate human-like reasoning. Such findings validate the effectiveness of integrating intent into ToM reasoning.

\begin{table}[htbp]
\centering
\footnotesize
\caption{The results from ROUGE-L, BLEURT, BERTScore, and MoverScore, obtained across various LLMs, were generated with prompt tuning based on users' emotion, intent, and sentiment of posts. The highest scores for each model are highlighted in \textbf{bold}, and the second highest with \underline{underline}.}
\label{tab:Final-results}
\begin{tabular}{l c c c }
\hline
\toprule
 & Zephyr-7B & Llama2-Chat-13B & GPT-4  \\
\hline
\rowcolor{gray!20}
\multicolumn{4}{c}{\textbf{ Sentiment}} \\
\hline
ROUGE-L & 0.244 & \underline{0.250} & \textbf{0.286} \\ 
BLEURT & 0.391 & \underline{0.401} & \textbf{0.428} \\ 
BERTScore &  0.586 & \underline{0.592} & \textbf{0.599}\\
MoverScore & \underline{0.310} & 0.303 & \textbf{0.316}\\
\hline
\rowcolor{gray!20}
\multicolumn{4}{c}{\textbf{Emotion}} \\
\hline
ROUGE-L & 0.284 & \underline{0.286} & \textbf{0.291}  \\ 
BLEURT & 0.440 & \underline{0.441} & \textbf{0.456}\\ 
BERTScore & 0.621 & \underline{0.622} & \textbf{0.627} \\
MoverScore & 0.342 & \underline{0.346} & \textbf{0.350} \\
\hline
\rowcolor{gray!20}
\multicolumn{4}{c}{\textbf{Intention}} \\
\hline
ROUGE-L & \underline{0.299} & 0.291 & \textbf{0.300} \\ 
BLEURT & 0.442 & \underline{0.445} & \textbf{0.452} \\ 
BERTScore &  0.629 & \underline{0.630} & \textbf{0.631} \\
MoverScore & 0.357 & \underline{0.360} & \textbf{0.368} \\
\hline
\rowcolor{gray!20}
\multicolumn{4}{c}{\textbf{Emotion + Sentiment + Intention}} \\
\hline
ROUGE-L & \underline{0.312} & 0.299 & \textbf{0.357}  \\ 
BLEURT & 0.460 & \underline{0.463} & \textbf{0.480} \\ 
BERTScore & 0.641 & \underline{0.647} & \textbf{0.661} \\
MoverScore & 0.381 & \underline{0.385} & \textbf{0.402}\\
\hline
\toprule
\end{tabular}
\end{table}

\subsubsection{\textbf{LLM reasoning with a focus on ``Sentiment + Emotion + Intention''}:}

In the combined tuning scenario, we use the prompt template from Section §~\ref{sec:Human} to incorporate sentiment, emotion, and intention into reasoning. This comprehensive prompt allows for a thorough exploration of how these intertwined mental states affect model performance. 

The scores evaluating the reasoning capabilities of the models are shown in Table~\ref{tab:Final-results}. Based on these results, GPT-4 demonstrates superior performance across all metrics. It achieves a ROUGE-L score of 0.327 and a BLEURT score of 0.472, showcasing its ability to incorporate and reason based on mental states. Llama2-Chat-13B often secures the second-highest scores, including a BERTScore of 0.647 and a MoverScore of 0.385. These scores reflect its ability to adapt to the complexities of considering emotion, sentiment, and intention simultaneously. Though it does not outperform GPT-4, Llama2-Chat-13B demonstrates capabilities in generating reasoning responses.

\begin{table}[htbp]
\centering
\footnotesize
\caption{Delta improvement in reasoning capabilities of LLMs through prompt tuning based on posts' sentiment, emotion, and intention.}
\label{tab:Delta}
\begin{tabular}{l c c c c }
\hline
\toprule
 & Zephyr-7B & Llama2-Chat-13B & GPT-4  \\
\hline
ROUGE-L & $\uparrow 7.29\%$ & $\uparrow 5.51\%$  & $\uparrow 7.88\%$ \\ 
BLEURT &  $\uparrow 6.30\%$ & $\uparrow 6.73\%$ &  $\uparrow 6.94\%$\\ 
BERTScore & $\uparrow 5.91\%$ & $\uparrow 6.29\%$ & $\uparrow 6.83\%$ \\
MoverScore & $\uparrow 8.14\%$ & $\uparrow 8.45\%$ & $\uparrow 9.27\%$ \\
\hline
\toprule
\end{tabular}
\end{table}

For a deeper analysis of the performance enhancements from incorporating a combination of emotion, sentiment, and intention into LLM reasoning prompts, we computed the delta changes between the combined format and the baseline results in Table~\ref{tab:results-Vanilla}. The delta results, displayed in Table~\ref{tab:Delta}, illustrate performance improvements across all evaluated metrics for each model.

In the ROUGE-L metric, all models showed notable improvements when prompted with combined mental states. GPT-4 led with a 7.88\% increase, followed by Zephyr-7B with 7.29\% and Llama2-Chat-13B with 5.51\%. For the BLEURT metric, GPT-4 again led with a 6.94\% improvement, indicating its ability to generate semantically rich text. Llama2-Chat-13B improved by 6.73\%, and Zephyr-7B by 6.30\%. In BERTScore, GPT-4 saw a 6.83\% rise, Llama2-Chat-13B improved by 6.29\%, and Zephyr-7B by 5.91\%. MoverScore showed the most pronounced improvements, with GPT-4 increasing by 9.27\%, Llama2-Chat-13B by 8.45\%, and Zephyr-7B by 8.14\%. These gains highlight the models' enhanced ability to capture fine-grained details of mental state information in the prompt template, which is crucial for advanced language understanding.

The analysis clearly shows that integrating emotion, sentiment, and intention into LLM prompts improves model performance, as evidenced by the human evaluation (Section §~\ref{sec:Human}). This integration not only improves the overall quality of the generated text but also aligns it more closely with human cognitive and emotional processing, making the outputs more natural. The consistent improvements across different models and metrics validate the effectiveness of prompt tuning based on mental states in advancing LLMs' capabilities in natural language understanding and reasoning.

\section{Discussion} \label{Discussion}
In this section, we will describe the aforementioned results in the reasoning capabilities of LLMs to address the research questions in this study.

\subsection{LLMs' capability in zero-shot reasoning with open-ended questions (RQ1)}

The analysis of annotations by human evaluators indicates that the reasoning responses to open-ended questions generated by LLMs do not meet high-quality standards, as detailed in Table~\ref{tab:Labeling-result}. This finding underscores a critical challenge in the field of artificial intelligence, specifically in the development and training of LLMs. Human evaluators consistently report that responses from these models, while often structurally sound and linguistically coherent, lack the depth, nuance, and contextual awareness inherent in human reasoning~\cite{hao-etal-2023-reasoning, wang-etal-2023-chatgpt-defend, gandhi2024understanding}. 

Moreover, as demonstrated in Table~\ref{tab:results-Vanilla}, the quantitative metrics used to evaluate model performance corroborate these observations by empirically highlighting the existing gaps, which align with the results of the human evaluation. These metrics, which assess aspects ranging from lexical similarity to semantic coherence, consistently reveal that despite their sophistication, LLMs still fall short of the natural reasoning processes humans use in responding to open-ended questions.

In conclusion, although LLMs are capable of mimicking human-like text generation, they still cannot fully replicate human-like ToM reasoning in open-ended questions. These results do not align with those of~\citet{kosinski2023theory, bubeck2023sparks}, who suggested that modern LLMs reached high ToM reasoning ability. We contend that these assertions are overly broad, derived from a limited focus on one aspect of ToM and based on a minimal set of examples. Furthermore, their experiments were conducted using multiple-choice and short-answer questions, whereas we explored LLMs' ToM abilities with open-ended questions, which more closely mirror real-life scenarios. In accordance with~\cite{sap-etal-2022-neural, kim-etal-2023-fantom, wu-etal-2023-hi, sclar-etal-2023-minding, strickland2023ai}, our analysis indicates that even the most advanced models are inadequate in ToM reasoning.

\subsection{LLM alignment with human ToM reasoning (RQ2)}

Based on the results presented in Table~\ref{tab:results-Vanilla} and \ref{tab:Final-results}, our analysis indicates that there is not a substantial alignment between human and LLM ToM reasoning capabilities when addressing open-ended questions. While LLMs mimic some aspects of human reasoning, they still fall short of fully replicating the depth and nuance inherent in human cognitive processes~\cite{hagendorff2023human, wang2024can, kim-etal-2023-fantom}. This suggests that despite advancements in LLM capabilities, significant discrepancies remain, underscoring the ongoing challenge of achieving a true equivalence in reasoning between humans and LLMs for open-ended questions.

\subsection{The impact of a questioner's mental state on LLM performance in ToM reasoning (RQ3)}

The integration of emotional and intentional information into LLMs has enhanced their capacity for reasoning, refining their effectiveness not only in multiple-choice and short-answer queries~\cite{gandhi2024understanding, kim-etal-2023-fantom} but also in open-ended questions. As shown in Table~\ref{tab:Final-results} and ~\ref{tab:Labeling-result}, by integrating mental state information, LLMs are able to generate higher-quality reasoning responses.  

This improvement is evident in enhanced performance metrics, showing that LLMs are increasingly able to process complex emotional and intentional information. This development is essential for ToM reasoning, enabling the models to handle open-ended queries with a depth similar to human interactions.

Despite advancements, challenges persist in achieving human-like ToM reasoning with LLMs. Prompt tuning has improved LLMs' ability to generate refined reasoning responses, yet it has not fully closed the gap between model outputs and the nuanced responses typical of human reasoning. This ongoing disparity is primarily due to the inherent subjectivity in reasoning, which allows for a wide range of valid responses influenced by individual perspectives and details~\cite{jsang2018subjective, ermer2006theory}. To minimize subjectivity and boost evaluation reliability, we analyzed five different answers per query, covering diverse human reasoning and reducing biases. Despite this, subjectivity still impacts the results, likely explaining the gaps in LLM responses compared to human reasoning. Additionally, disparities inherent within the LLM models may also contribute to this shortfall.

\section{Conclusion} \label{Conslusion}

In the evolving landscape of AI, the pursuit of integrating ToM reasoning into LLMs poses opportunities alongside challenges. This study explores how well LLMs' ToM reasoning aligns with human reasoning in handling open-ended questions. Despite advancements in LLMs, our findings reveal a clear gap in their ToM reasoning capabilities for open-ended questions.

Our comprehensive evaluation, utilizing data from Reddit's ChangeMyView subreddit, has elucidated the limitations of LLMs. Despite the proficiency of models such as Zephyr-7B, Llama2-Chat-13B, and GPT-4 in structured tasks like summarization~\cite{ahmed2022few, kolagar2024aligning, tang2023evaluating}, question answering~\cite{abbasiantaeb2024let, tan2023can}, and language translation~\cite{lu-lin-2023-characterised, kocmi2023findings}, they exhibit deficiencies in ToM reasoning with open-ended questions. Our findings indicate that these LLMs fall short of producing high-quality reasoning that aligns with human capabilities in such contexts. The significant discrepancy between human and LLM reasoning in ToM can be partly attributed to the inherent subjectivity involved in ToM reasoning ~\cite{ermer2006theory}. To mitigate this subjectivity, we analyzed five different reasoning strategies for each open-ended question. However, despite this methodological diversification, our results reveal that LLM outputs still diverge considerably from human reasoning, underscoring a critical limitation in their current ToM capabilities.

Our study confirmed that integrating human intentions and emotions through prompt tuning enhances LLMs' ToM reasoning abilities. By embedding individual emotions, intentions, and sentiments within the prompts, we improved the LLMs' ability to generate responses that more closely mirror human ToM reasoning, as demonstrated in~\cite{gandhi2024understanding}. Despite these advancements, the reasoning results still fall short of fully aligning with human-level reasoning in open-ended quetions.

\section{Limitations and Future Work} \label{limitation}

This study shows that incorporating mental states through prompt tuning improves ToM reasoning in LLMs of open-ended questions, yet it faces limitations. The first limitation involves subjectivity. Although we consider five different reasoning answers to cover a diverse spectrum and minimize subjectivity, subjectivity remains a potential factor in the differences observed between LLMs and human reasoning. Moreover, while integrating individual intentions and emotions has proven to boost LLMs' ToM reasoning capabilities, it remains uncertain whether LLMs can inherently account for these mental states in their reasoning processes without explicit prompt tuning. In future work, we will investigate whether LLMs inherently account for mental states in their ToM reasoning without explicit integration.

Another limitation of this study stems from its data source; we gathered open-ended questions and responses from the Reddit community. While these responses are tailored to Reddit's unique cultural context, they have been generalized to reflect the broader global community in our analysis. In this categorization, Reddit users are simply referred to as `humans'. Moreover, it is crucial to recognize that our data is exclusively from English language interactions, potentially limiting the relevance of our findings to LLM behavior across different languages.

\newpage

\bibliography{Ref}
\bibliographystyle{ACM-Reference-Format}

\end{document}